\documentclass[iicol]{sn-jnl}

\usepackage[numbers]{natbib}
\usepackage{tabularx}
\usepackage{booktabs}
\usepackage{tablefootnote}
\usepackage{makecell}
\usepackage{adjustbox}
\usepackage{lmodern}

\jyear{2023}%

\theoremstyle{thmstyleone}%

\theoremstyle{thmstyletwo}%

\theoremstyle{thmstylethree}%

\newcommand{\changed}[1]{#1}
\begin{document}

\title[A Unified Representation Framework for the Evaluation of
Optical Music Recognition Systems]{\changed{A Unified Representation Framework for the Evaluation of Optical Music Recognition Systems}}


\author*[1]{\fnm{Pau} \sur{Torras}}\email{ptorras@cvc.uab.cat}
\author[1]{\fnm{Sanket} \sur{Biswas}}\email{sbiswas@cvc.uab.cat}
\author[1]{\fnm{Alicia} \sur{Fornés}}\email{afornes@cvc.uab.es}

\affil*[1]{\orgdiv{Computer Vision Center, Computer Science Department}, \orgname{Universitat Autònoma de Barcelona}, \orgaddress{\city{Cerdanyola del Vallès}, \postcode{08193}, \country{Spain}}}


\abstract{Modern-day Optical Music Recognition (OMR) is a fairly fragmented field. Most OMR approaches use datasets that are independent and incompatible between each other, making it difficult to both combine them and compare recognition systems built upon them. In this paper we identify the need of a common music representation language and propose the Music Tree Notation (MTN) format, with the idea to construct a common endpoint for OMR research that allows coordination, reuse of technology and fair evaluation of community efforts. This format represents music as a set of primitives that group together into higher-abstraction nodes, a compromise between the expression of fully graph-based and sequential notation formats. We have also developed a specific set of OMR metrics and a typeset score dataset as a proof of concept of this idea.}




\maketitle

\section{Introduction} \label{sec:intro}
%

Written music has been part of the cultural heritage of humankind for many centuries.
From neumes as old as a thousand years to current-time Western notation scores, people have built methods to make this most ephemeral of arts persistent over time. Rivers of ink have been imprinted on libraries worth of paper to preserve music along the ages, and much of it has indeed survived until today.

It is therefore unsurprising that, given the sheer volume of notably interesting
(and oftentimes, forgotten) pieces of music that have been endowed to our current
generations, scholars have turned their attention to computers to aid them in the
endeavour of preserving and analysing them. The task that concerns us in this work is
Optical Music Recognition (OMR), which is fundamental step for
the computational analysis of written scores; converting images or scans of music into
a defined format a computer can process \cite{calvo-zaragoza_understanding_2021}.

Modern-day OMR has been greatly enriched with the advent of the powerful deep-learning
techniques developed during the last decade \cite
{calvo-zaragoza_understanding_2021,rebelo_optical_2012}. Nevertheless, the field itself
remains fragmented, with few researchers fully devoted to it and each of them
developing their unique point of view and methodology  \cite
{pacha_advancing_2018,calvo-zaragoza_understanding_2021}. This is particularly evident
when analysing the available datasets \cite{pacha_omr_2017}, as can be seen in Table \ref{tab:datasets}; most of them are
restricted to specific steps or approaches \cite
{parada-cabaleiro_seils_2017,tuggener_deepscores_2018,calvo-zaragoza_camera-primus_2018}
and almost none of them are compatible with each other (the most notable exception
being the DoReMi dataset introduced in 2021 \cite{shatri_doremi_2021}, which incorporates its ground truth in multiple formats).

\begin{figure*}
    \centering
    \includegraphics[width=\textwidth]{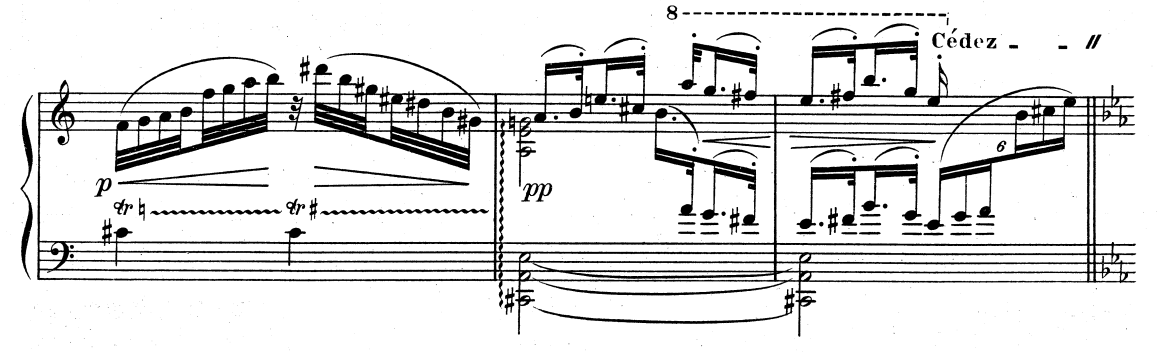}
    \caption{Fragment of a scan of a 1910 edition of Claude Debussy's \textit{La Danse
     de Puck} from the first volume of his \textit{Préludes}. In this fragment, many of
     the reasons why modelling OMR scores is difficult can be seen -- a two-staff
     system with voices that move along both of them, certain areas of
     the score accumulating many small symbols, many independent voices sounding in unison,
     elements spanning multiple measures, chords, etc.}
    \label{fig:debussy_score}
\end{figure*}

Another point of disagreement among OMR
researchers is the matter of the evaluation of models \cite
{hajic_muscima_2017,hajic_jr_case_2018,mengarelli_omr_2019,calvo-zaragoza_understanding_2021},
which is but the visible consequence of many long-standing issues -- some related to
the inherent difficulty of the field (for an example, see Figure \ref
{fig:debussy_score}), but others related to the diversity of plausible approaches to
music recognition.

Evaluation of OMR models is currently performed on a per-methodology basis \cite
{calvo-zaragoza_understanding_2021,rebelo_optical_2012}. If one considers the goal of
OMR to produce a computational representation of music -- whichever that final
representation might be \footnote{Byrd \textit{et al.} \cite{byrd_music_2003} collected
a very extensive and detailed set of requirements for a music notation system that is
suitable for both score engraving and OMR, but to the best of our knowledge there is no
recognition system that formally or explicitly applies them.} -- one would expect the
metrics to obey a criterion of either some aspect of music understanding or quantifying
the correctness of the final representation. Instead, most metrics revolve around
measuring the fidelity of intermediate representations devoid of musical meaning
(e.g. sequences of tokens or bounding boxes),
disregarding the full reconstruction of music scores altogether. Moreover, the
benchmarks these evaluation metrics are computed upon are rarely widespread, with each
methodology using their own. Commercial OMR systems have also been very vague about
the specifics of their capacity.

With the aim of setting a step forward towards unifying the efforts of the community, we propose a way to set a shared framework for OMR. We believe \textbf{the first step is to set a target final representation that accommodates a significant amount of use cases within Common Western Music Notation (CWMN)}. This is the music notation system that has been employed in Europe from the early 18th century until today.

Focusing on CWMN specifically instead of the broad range of western notations related to it is not an arbitrary choice. While CWMN did indeed evolve from Mensural notation, which was employed from the 13th to the 17th century, and consequentially shares a considerable amount of graphical constructs, this system has semantic concepts that make them quite distinct. For instance, the concept of beat as we know it today was developed during the first half of the 18th century; before that, the rules that described the relation of note types with their divisions implied different abstractions and note types that are not compatible with those of modern notation. Moreover, there are constructs such as ligatures that have very unique properties but have disappeared from CWMN.  In this regard, given that each notation system has a distinct nature and given the necessity for OMR systems in CWMN, we believe it is best to commit to this specific notation alone.

Once this target representation is chosen, a way of easily working with it has to be developed. Currently used encoding formats are quite
exhaustive, verbose and option-filled, which allows expressing the same score in a
variety of ways -- something that makes them impractical for both recognition and direct comparison.
Thus, we propose an intermediate representation that translates directly to the desired
output and offers analysis, conversion and evaluation tools regardless of the underlying
approach. We propose doing so through a notation
based on the idea of building a tree structure; leaves are the objects in the music score
and intermediate nodes are their combination. The idea is exploiting the structural
advantages of graph-based approaches over sequence-based ones while restricting the algorithmic complexity of working with arbitrary graphs. Incorporating a shared notation format is also an opportunity for unifying all existing datasets so that they can be both combined and compared on equal grounds, regardless of the methodology they were originally designed to work with.

The contribution of this work can be summarised by the following claims:
\begin{itemize}
    
    \item We try to bridge the gap between the different benchmark suites in OMR
     literature with a universal tree-based notation format designed to represent
     musical scores at the graphical level\footnote{Repository of the project: \url{https://github.com/CVC-DAG/comref-converter}}.
    \item We also present an evaluation toolkit which aims towards unify existing benchmark OMR tasks for fairer comparison.
    \item We have produced a typeset dataset using public domain works with permissive
     licenses \footnote{Link to the dataset \url{https://datasets.cvc.uab.cat/comref/comref.zip}}.
\end{itemize}

This paper is structured as follows. In section \ref{sec:terminology} some definitions
for recurrent topics in the article are presented. Then, section \ref{sec:rationale}
justifies the necessity for an OMR evaluation framework and debates about how that
framework should be structured. In section \ref{sec:format}, our proposed notation format
is presented and described in detail, whereas in section \ref{sec:metrics} we describe
the evaluation metrics to be employed with it. In section \ref{sec:data} we introduce the proof-of-concept dataset on typeset scores. Finally, section \ref{sec:conclusion} closes the
paper with some final remarks and future work.
\section{Terminology} \label{sec:terminology}
In this section we shall address some specific recurrent terms for the sake of clarity and cohesiveness of the text.
Throughout this paper we shall adhere to the nomenclature and definitions found in Calvo-Zaragoza \textit{et al.}'s \cite{calvo-zaragoza_understanding_2021}. 

By \emph{structured output} we refer to an output encoding format that allows the replication of the input score exhaustively and unambiguously. This is in contrast to formats such as MIDI, that only encode certain aspects of music (in this case, playback information).

Another relevant point is the definition of an \emph{end-to-end} OMR model. Modern day literature employs this term frequently, but it must be noted that it is used with a slightly different meaning than the term itself actually conveys. The most widely recognised definition of the OMR pipeline is perhaps that of the review authored by Rebelo \textit{et al.} \cite{rebelo_optical_2012}, which states four distinct steps:
\begin{enumerate}
    \item Image preprocessing.
    \item Recognition of Musical Symbols.
    \item Reconstruction of Musical Information.
    \item Construction of a Musical Notation model.
\end{enumerate}
Most end-to-end approaches do not actually perform all four steps; rather, these models stop at step 3 with a notation that encodes most of the score's information directly or indirectly, but cannot be directly engraved (sometimes requiring an additional processing step before all music semantics can be successfully generated). Whenever we address end-to-end approaches, we refer to these.
In this work we mostly address \emph{offline} use cases of OMR -- that is, recognition of raster images of scores. Nevertheless, we believe some of our ideas are equally applicable to online use cases \cite{calvo-zaragoza_recognition_2014}.
\section{The Necessity for an OMR Framework} \label{sec:rationale}
In this section we comment on some of the main issues of the field of OMR and provide arguments as to why having a unified OMR framework is useful and necessary.
\subsection{OMR Needs a Standardised Output}
Most OMR research today does not tackle the problem of converting a model's output to a
symbolic representation that can be engraved into a proper score. In their review of
the field in 2021, Calvo-Zaragoza \textit{et al.} \cite
{calvo-zaragoza_understanding_2021} go as far as claiming that no research OMR system
that they know of is actually capable of performing such a conversion\footnote
{There are however commercial OMR systems\cite{bitteur_audiveris_2004} that can perform
this step.}, and since its publication we have only acknowledged one system that
can \cite{shishido_listen_2021}. One of their main justifications for this is the fact
that most structured music engraving formats are either not mature enough to cover all
possible use cases or impractical for OMR tasks due to their complexity.

There are currently two main choices for music score exchange formats: 
MusicXML \cite{good_musicxml_2001}, devised by Recordare and now managed by a W3C committee and the Music Encoding Initiative (MEI), designed by Perry Roland \cite{roland_music_2002}. Other formats such as Lilypond \cite{the_lilypond_developement_team_lilypond_2014} (an open-source text-based music engraving software) or Humdrum **kern \cite{HumdrumToolkitComputational} (a notation format for the automatic musical analysis tools in the Humdrum suite) have been used for OMR as they allow engraving scores very succinctly. Nevertheless, these formats are not designed to represent scores precisely and unambiguously -- beaming groups are not explicitly modelled, multiple representations of the same score through permutations are allowed (although **kern has been extended to fix this \cite{rios-vila_end--end_2023}), ornaments, expression signs, dynamics and other non-structural elements are only partially supported, etc. As a consequence, these formats are not suited to engrave exact replicas of any music score.

MusicXML and MEI are somewhat similar in their base design ideas, with the former being
more widely used within music score repository platforms and engraving software and the
latter in archival settings. Thus, we believe it is important
to make all OMR applications interoperable with these formats by design. Doing so makes it
easier to collect data for OMR applications and also simplifies deploying OMR systems. 
Having a well-defined target representation has the additional side-effect of fixing the requirements 
for what is expected from an OMR system, encouraging joining efforts within the community.

Since working with these formats directly can be difficult, it would be interesting
to have a different intermediate representation that acknowledges the requirements of
OMR but can convert to other formats when required.
\subsection{Trade-offs for Structured Representations of Music for OMR}
To this day, there is no straightforward answer as to what is the best basic data
structure to represent music for OMR purposes. There are various approaches being
explored in the community, each having their own set of trade-offs.

The most flexible -- and perhaps the most logical -- computational representation of
written music is most likely a graph, since it allows expressing
arbitrary relationships between primitives and symbols. There exists a very widely
known notation format based on this idea called Music Notation Graph (MuNG) introduced
for the MUSCIMA++ Dataset \cite{hajic_muscima_2017}. Some works have indeed succeeded in being
able to reconstruct music structure from these graphs, such as Baró \textit{et al.}'s
Musigraph \cite{baro_musigraph_2022} or Garrido-Muñoz \textit{et al.}'s Convolutional
Recurrent Neural Networks \cite{garrido-munoz_holistic_2022}. Nevertheless, to the best
of our knowledge, there is no straightforward way of converting this graph output back
into a structured encoding. Moreover, the cost of evaluation on graphs is high due to
the complexity of matching them.

At the other end of the spectrum lies the sequential music model, currently employed in
some form or another by most end-to-end OMR approaches \cite{calvo-zaragoza_end--end_2017,calvo-zaragoza_end--end_2018,baro_optical_2019,alfaro-contreras_exploiting_2021,alfaro-contreras_decoupling_2022,rios-vila_use_2022,baro_optical_2019,baro_handwritten_2020,torras_integration_2021,wen_sequence--sequence_2022,edirisooriyaEmpiricalEvaluationEndtoEnd2021,rios-vila_end--end_2023}.
These methods impose a relationship between the horizontal axis of the image input and
the playback time of the score. However, this forces either having to model music time semantics when inferring the structure of the score or introducing simplifications to music notation. The latter is the most common, with information such as stem directions or beaming groups being abstracted away to varying degrees.
\subsection{Annotated Data is Scarce}
\begin{table*}[ht]
\caption{Description of annotation formats available for OMR datasets designed for final output production (i.e. they are designed to reconstruct a full score in some form). Columns from left to right: the publication that spawned the dataset, the name it is usually referred to as, what type of music samples it contains and whether it contains bounding box annotations, pixel-mask annotations, notation graph (MuNG) among symbols, string-based output annotations and a final structured output (MusicXML or MEI).}
\label{tab:datasets}
\begin{tabularx}{\textwidth}{@{}p{2cm}p{2.5cm}p{3cm}XXXXX@{}}
\toprule
\textbf{Publication}                                        & \textbf{Dataset Name} & \textbf{Dataset Type}                         & \textbf{Boxes?} & \textbf{Masks?} & \textbf{Graph?} & \textbf{String?} & \textbf{Struct?} \\ \midrule
\cite{hajic_muscima_2017}                                   & MUSCIMA++             & \makecell{Handwritten \\ Contemporary}        & \checkmark      &                 & \checkmark      &                  &                  \\
\cite{tuggener_deepscores_2018, tuggener_deepscoresv2_2020} & DeepScores            & \makecell{Typeset}                            & \checkmark      & \checkmark      &                 &                  &                  \\
\cite{tuggener_real_2023}                                   & RealScores            & \makecell{Scanned}                            & \checkmark      & \checkmark      &                 &                  &                  \\
\cite{shatri_doremi_2021}                                   & DoReMi                & \makecell{Typeset}                            & \checkmark      & \checkmark      & \checkmark      & \checkmark       & \checkmark       \\
\cite{parada-cabaleiro_seils_2017}                          & SEILS                 & \makecell{Typeset \\ Scanned \\ Mensural}     &                 &                 &                 & \checkmark       & \checkmark       \\
\cite{baro_handwritten_2020}                                & Baró Synthetic        & \makecell{Typeset}                            &                 &                 &                 & \checkmark       &                  \\
\cite{baro_handwritten_2020}                                & Pau Llinás            & \makecell{Handwritten \\ Historical}          &                 &                 &                 & \checkmark       &                  \\
\cite{calvo-zaragoza_end--end_2018}                         & PRIMuS                & \makecell{Typeset}                            &                 &                 &                 & \checkmark       & \checkmark       \\
\cite{calvo-zaragoza_camera-primus_2018}                    & Camera PRIMuS         & \makecell{Typeset}                            &                 &                 &                 & \checkmark       & \checkmark       \\ \bottomrule
\end{tabularx}
\end{table*}
One of the longest-standing problems in OMR is the lack of reliably annotated
data for contexts other than recognition of typeset scores. Even in that case,
the datasets are usually tailored to specific recognition techniques and rarely
incorporate structured music sources (see Table \ref{tab:datasets} for more detail). Most
datasets are incompatible between each other as a result.

In this sense, it would be very positive that all datasets provide an additional
structured source in a common format in a way similar to \cite{shatri_doremi_2021},
so that the original data can be used in contexts outside the ones initially intended.
This would help develop better systems overall by having the possibility of ensembling datasets
in those domains where there are many of them available.

However, there are some domains for which there is barely any data available.
An example of this are historical CWMN, written from between the 17th and 20th centuries.
As can be seen in Table \ref{tab:datasets}, only Baró \textit{et al.}'s
\cite{baro_handwritten_2020} Pau Llinás dataset is dedicated to music from this period. This is caused for various reasons:
\begin{itemize}
    \item The only realistic way of annotating a corpus of handwritten music from scratch given the tools existing today is doing so manually, which is extremely time-consuming.
    \item Crowdsourcing annotation of music scores is difficult because it requires trained individuals. Some music scores in this period are notoriously difficult to read.
\end{itemize}

Making it possible to incorporate structured sources opens up the possibility of using existing
public domain transcriptions works of widely known authors and trying to synchronise them to the
original public manuscripts (e.g. Bach \cite{noauthor_bach_nodate}
or Beethoven \cite{noauthor_beethoven-haus_nodate}) at a significantly lower cost.
\subsection{Evaluation of OMR Models}
Evaluating OMR models fairly is important in order to be able to understand the
strengths and weaknesses of each system against the rest and being able to dissect the
effect of design decisions on the final output.

Unfortunately, most metrics used by the OMR community in modern day are fairly
domain-specific and rarely take into account a final representation of music
\cite{mengarelli_omr_2019}. Authors
that address OMR as an object detection problem report object detection metrics,
whereas authors that address OMR as an image-to-sequence task rely on sequence edit
distances. Consequently, there is no realistic way of comparing them since the
semantics of the mistakes of the system are lost and the underlying notations are
different. The few authors that do report musically-aware metrics restrict themselves
to per-note pitch and/or time accuracy \cite{baro_optical_2019,huang_state---art_2019}.

In this direction, there is also debate regarding the specific aspects of music that
should be measured. Hajič jr. \cite{hajic_jr_case_2018} argues that the best metric is
one that summarises the entire notation system as one single value, reminiscent of the
Symbol Error Rate (SER) metric for strings. We believe that, while such a metric is a
good indicator of overall performance (and is definitely useful for benchmarking), it
must not be considered on its own, as definite conclusions about the underlying models
are very hard to make. Our belief is that there should be an array of metrics that are
oriented at different aspects of the score that should be logged together, increasing
the coverage of the system. Both cases only make sense under the assumption that
comparisons are made on a standardised set of objects and primitives.
\subsection{The Ideal Scope for Recognition}
In this final section we discuss the effect of tackling OMR at different levels of
detail.

Presently, there is no pre-defined scope to tackle the problem of OMR. Therefore
researchers can choose whether to implement recognition at measure \cite{baro_handwritten_2020,baro_musigraph_2022}, line \cite{calvo-zaragoza_camera-primus_2018,baro_optical_2019,wen_sequence--sequence_2022} or
page level \cite
{pacha_baseline_2018,rios-vila_end--end_2022,zhang_detector_2023}. While there are
practical advantages on each perspective, there is an arguing point that should
be brought to discussion given the current situation of OMR and Computer Vision as a
whole.

Current state-of-the-art object detection and recognition models struggle with both
small \cite{yu_1st_2020,pacha_challenge_2021,baro_musigraph_2022} and scale-variant
symbols \cite{zhang_multi-scale_2018}. Music scores are quintessential examples of both
phenomena: large symbols (slurs, ties, staves) coexist with very small ones
(dots, fingering numbers), and writers draw symbols at their preferred relative scale
without affecting their meaning -- something that is particularly true for noteheads
and stems -- while also writing semantically-distinct symbols that change based on
their size -- grace notes. Additionally, the number of objects present in the input image
also affects the recognition performance, particularly when objects accumulate in a limited
region of the image.

Considering this, it should be possible to tackle OMR from the smallest scope possible: 
the measure level. Addressing higher scopes can be achieved by composing smaller semantic units.
The downside is that this requires more effort to be put in the analysis of the layout of the scores, as all successive recognition steps depend on finding the measures reliably on the page. Nevertheless, it can be argued that it is an objectively simpler problem than recognition: staves and bar
lines are very prominent, very ubiquitous and can easily be identified
heuristically \cite{fujinaga_staff_2004,cardoso_staff_2009,fornes_2012_2013,bui_staff_2014,calvo-zaragoza_staff-line_2017,egozy_computer-assisted_2022},
as opposed to many of the other musical symbols.

\noindent\rule{\linewidth}{1pt}

In this section we have provided the core ideas for a common OMR notation format
and evaluation procedure. This shall be the logical foundation upon which our proposed
solution, described in the following section, is built.
\section{The Music Tree Notation Format} \label{sec:format}
As stated in Introduction and in concordance to the necessities discussed in
Section \ref{sec:rationale}, we have designed a notation format that
\begin{itemize}
    \item normalises the set of music primitives to be recognised,
    \item simplifies conversion to a final structured format,
    \item enables comparison of diverse OMR methods on equal grounds and
    \item facilitates the usage of non-OMR-specific data.
\end{itemize}
The base logic for the notation system is introduced in \ref{sub:rationale}. The
notation system is described in detail in \ref{sub:description}.
\begin{figure*}[ht]
    \centering
    \includegraphics[width=\textwidth]{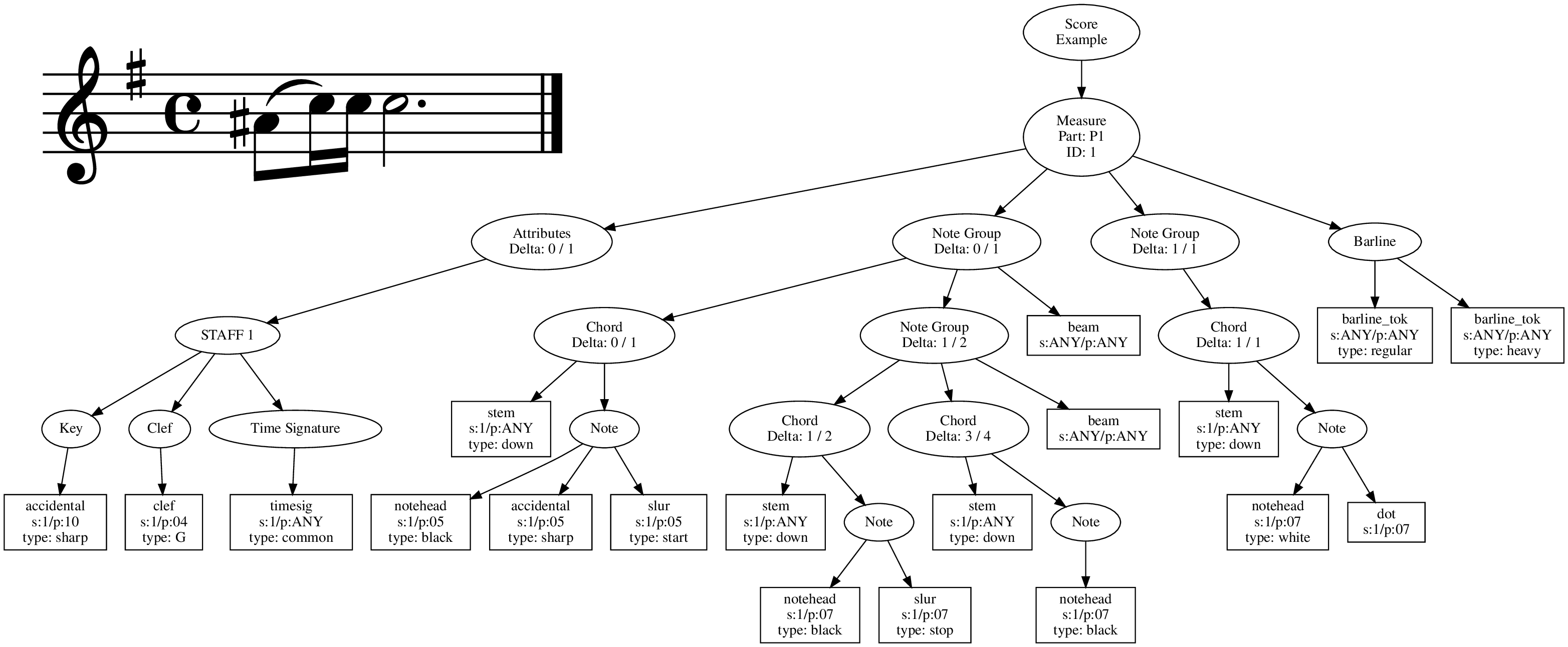}

    \caption{Example showing a fragment of a measure in which the annotation format for
     attributes and staff-modifying elements is shown. Rectangular nodes represent primitives
     as tokens and rounded nodes are abstract elements.}
    \label{fig:notation_format_attr}
\end{figure*}
\subsection{Rationale} \label{sub:rationale}
The basis to our proposal for an OMR framework for inference, score 
reconstruction and evaluation is
the Music Tree Notation format (MTN). We believe the final goal of structured OMR is
the reconstruction of the score at the visual domain. The core idea of this format is
therefore to build a notation that exclusively models relationships between graphical
symbols and defers inference of music semantics until a later stage. Only those
high-level music concepts that are strictly required to reconstruct the score
unambiguously are kept if and only if there is a direct graphical cue that allows 
straightforward inference.

\textbf{Score reconstruction should be possible regardless of the scope of recognition of the underlying system} -- page, line or measure. Nevertheless, a minimal semantic unit is still
required in order to reconstruct music semantics afterwards and facilitate alignment of
predictions and their groundtruth. Exploiting the assumption OMR is performed on
the visual domain and therefore application of semantic-altering objects is irrelevant,
we propose using the measure as the minimal unit of music on which to evaluate.
Moreover, reconstruction should be possible regardless of the origin of the transcription.
The format must not require localisation of symbols -- only their relationships -- in order to
accommodate the use of end-to-end or pipelined techniques interchangeably.

\textbf{MTN should act as a final target for OMR}, through which to convert the score 
into any other high-level notation system. Employing a tree structure simplifies
developing converters to new formats, which can be implemented by writing a tree 
traversal algorithm at linear complexity.
This has the potential of facilitating OMR systems from
research being deployed into real use-cases as well as enabling any source of
structured data to be used in OMR applications. Therefore, MTN draws heavy
inspiration from both MEI and MusicXML, but ``normalises'' their structure in such a
way that every score can only be expressed in one way. This also allows implementation
many of the already-studied evaluation requirement ideas from other authors,
particularly Hajič jr. \cite{hajic_jr_further_2016} and Byrd \cite{byrd_towards_2015}.

Finally, music notation is very rich and tends to bend its own rules quite heavily. \textbf{The
goal for this notation and evaluation system is to accommodate a most significant
subset of CWMN}, but covering every single exception proves to be a daunting task. Byrd
\cite{byrd_towards_2015} presents some specific cases such as a spanning slur with seven
inflection points that is rather complicated to model in the logical domain.
Therefore, our rationale is that support for these should not be a priority \footnote{This specific change could potentially be modelled in MTN using the tools already in the format either by providing the bounding box of the slur (which would be rather imprecise) or its pixel mask associated to its ID in the tree.}. We
believe it is better to have some solid tools on a core subset of music rather than
trying to cover every possible case with highly complex and fragile ones as this makes
adoption of the format much more difficult. In this regard, the format is designed so it is very easy to add support for new types of music primitives or add new sources of information.
\subsection{Description of the Format} \label{sub:description}
A graphical representation of a simple measure engraved in MTN can be seen in Figure \ref{fig:notation_format_attr}.
The core element of this format is the Musical Primitive, a concept that is quite
widespread in the OMR literature
\cite{baro_towards_2016,baro_handwritten_2020,calvo-zaragoza_camera-primus_2018,tuggener_deepscores_2018}
and can be defined as any of the independent structural elements that may or may not
be combined together to form a semantic unit in the music score. The set of musical
primitives includes all graphical elements in a score that are self-contained and
require no other symbols to convey meaning (this includes rests, clefs or time
signature symbols), the set of graphical elements that compose notes (noteheads, stems,
flags, dots, accidentals, etc.) and other miscellaneous elements such as numbers for compound time
signatures. Every primitive is given a unique work-level identifier.

These primitives associate together to form more abstract constructs. This is modelled
in MTN using a tree-like structure of higher-order elements resembling an Abstract
Syntax Tree (AST), which defines the set of dependencies among objects in the score.
This idea emulates parsing the contents of the score using a grammar, enabling the bulk
of tools and research on parsers, parser generators and AST analysis and processing to
be used in the context of music. Furthermore, it is a structure that can be modelled
very easily using an exchange format such as XML. The structure of the
format may be found in Table \ref{tab:structure}.

Tree-based music representations for evaluation of recognition has been explored before in \cite{foscarinDiffProcedureMusic2019}. In this paper they also nest tuplets within the hierarchy.
Nevertheless, this is not possible for any tuplet in MTN because they can extend beyond the scope of the current group and incorporate neighboring rests or other objects, which would break the tree assumption of the format. Instead, this is modelled using two start/end tokens.

\begin{table*}[ht]
    \caption{Definition of the main abstract elements within MTN and the overall structure of the format. Each of the lowest levels of the notation contain the various final tokens.
    For more information, refer to the implementation of the format.}
    \label{tab:structure}

    \begin{tabularx}{\textwidth}{@{}p{1.8cm}p{1.8cm}p{1.8cm}X@{}}
        \toprule
        \textbf{Top Level} & \textbf{Level 1} & \textbf{Level 2} & \textbf{Definition}                                                                   \\
        \midrule
        Attributes         &                  &                  & Any semantic change the score goes through.                                           \\
                           & Staff            &                  & Group of changes occuring on the same staff.                                          \\
                           &                  & Clef             & A change of clef in the score.                                                        \\
                           &                  & TimeSig          & A change of time signature in the score.                                              \\
                           &                  & Key              & A change of key in the score.                                                         \\
        \midrule
        Barline            &                  &                  & A measure separator and its attached modifiers.                                       \\
        \midrule
        Direction          &                  &                  & A single playback directive.                                                          \\
        \midrule
        Note Group         &                  &                  & Group of notes tied together by at least one beam.                                    \\
                           & Note Group       &                  & \textit{Note Groups can be nested, each owning one or more beams}           \\
                           & Chord            &                  & Set of notes in unison attached to the same stem, if it exists.             \\
                           &                  & Stem             & Stem direction information and beams or flags attached to it.                         \\
                           &                  & Note             & A notehead and its attached modifiers.                                                \\
        \midrule
        Rest               &                  &                  & Rest token and its attached modifiers.                                                \\

        \bottomrule
    \end{tabularx}
\end{table*}

There are some elements in music that break the tree-like structure
assumption. These are elements that connect multiple notes together outside their local
note group structure: slurs, ties, parentheses and tuplets, among others. Both MEI and
MusicXML acknowledge this limitation and circumvent it through the use of identifiers.
MTN is no different: it provides a unique starting and ending token for each side of
the object and gives both ends the same identifier.

In order to describe the position of MTN elements, two
magnitudes are used. Firstly, for every token a tuple of two integers denotes the staff the
element belongs to and its position within the staff. The position is denoted counting
the number of steps from the first ledger line below a staff. For those elements
without a specific position (such as rests or stems), a null value is used. Secondly,
for any object immediately below the class measure, an exact timing value is provided.
It is measured in fractions of a quarter note from the start of the measure itself.
This information is also provided for every chord in a note group even if this
information can be inferred for the sake of simplifying evaluation procedures.

\begin{figure*}
    \centering
    \includegraphics[width=\textwidth]{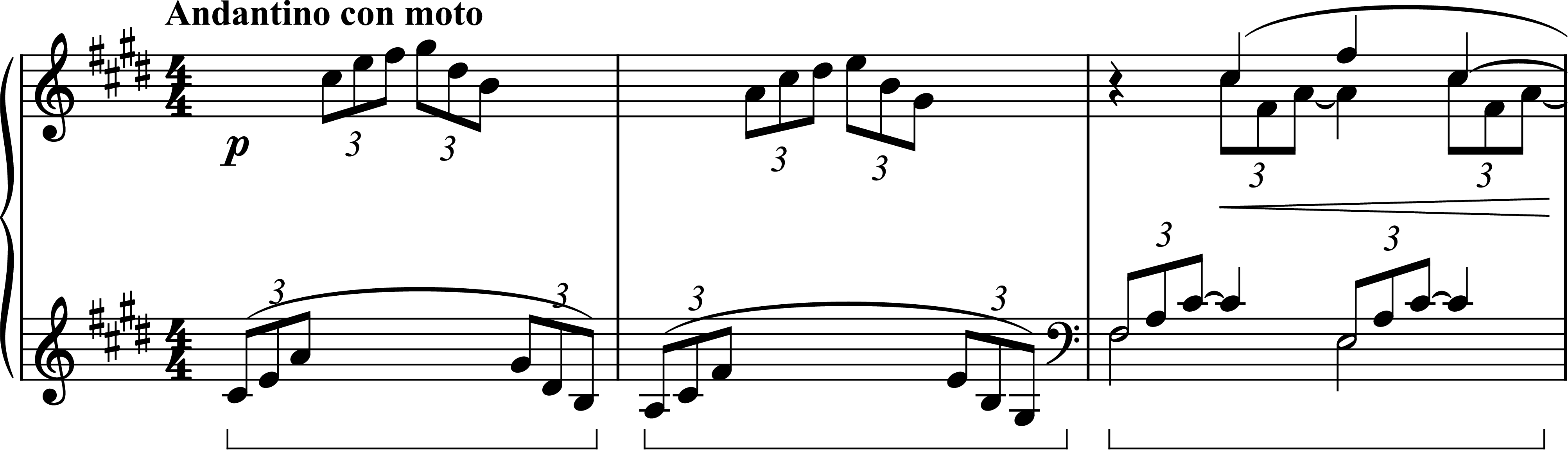}
    \caption{The first few measures on Claude Debussy's first Arabesque. The single voice
    moves along both staves and is played using both hands on a piano. If one was to separate
    both staves, it would be impossible to extract the exact timing where each group of notes
    is to be played. Another interesting aspect of this piece is that many editions use
    implicit triplets, something that only can be inferred contextually by the fact that eighth
    notes are on groups of three.}
    \label{fig:arabesques}
\end{figure*}

This musical information is necessary for two reasons. The only visual cue for aligning
elements in a polyphonic setup is that, at some point, additional voices have at least
one object that is perfectly aligned with one in the main voice. This is very hard to
model cleanly with anything other than a graph. Moreover, the main voice of a measure
may arbitrarily jump from staff to staff in a system, making timing generally
impossible to resolve on a per-staff basis, as can be seen in Figure \ref{fig:arabesques}. A small optimisation for sequence-based notation systems is that
only the first element in a group needs timing information, as all successive
chords' timings can be inferred from the note types and tuplets attached to them
(or their surrounding elements).

Doubly stemmed notes (such as in the last measure of the Arabesques example in Figure \ref{fig:arabesques})
can be interpreted as two notes playing at the same time. Therefore, they can be modelled
using two distinct groups without needing to break the tree-like structure.

Finally, to produce unambiguous scores, a reading order of sorts must be established.
We propose the following ordering criterion:
\begin{itemize}
    \item By starting time counting from the beginning of the measure.
    \item By top level class:
    \begin{enumerate}
        \item Attributes
        \item Directions
        \item Rests
        \item Note Groups
        \item Barlines
    \end{enumerate}
    \item By staff position: first objects on upper staves and lower positions within
     them.
    \item In case of Note Groups, by direction of the first stem: first stems looking
     upwards.
    \item For anything else, token alphabetical order. This also guarantees stability of
     the notation if new token types are added.
\end{itemize}
\subsection{Extensions} \label{sub:extensions}
The main goal for the notation is to address symbolic-level OMR. Nevertheless, many
practitioners use pipelined architectures where some partial results need to be
annotated (most notably, object bounding boxes). Providing these annotations with MTN
is straightforward: it is a matter of adding an ``annotations'' field with each token
identifier linked to the desired information. This also includes some other relevant
semantic cues such as voices, which we deem unnecessary for the task of score
reconstruction but might be useful for some other music information retrieval tasks. A
concern surrounding voice information is that its definition around music engraving
software is more akin to a layer in a graphic design suite; it is mostly used as a
way of circumventing engraving limitations, making voice information in files like
MusicXML rather arbitrary.

The presence of text in scores can also be treated separately from the rest of music
semantics. Since its interaction with the score is very situational and depends a lot
on the contents of the text itself, it is very difficult to devise a set of
relationships that might be useful for most scores. Therefore, the only text that is to
be represented in MTN are dynamic directives, which are common and well-defined. For
the rest of the text such as tempo indications or other textual remarks, we believe it
is best to append a set of text boxes in a similar way as bounding boxes are provided.

 The specification of the format provides ways of developing new extensions as new needs arise.  
\section{Evaluation Metrics} \label{sec:metrics}
We propose a set of evaluation metrics that both acknowledge the existence of multiple paradigms for OMR while also setting ways to compare any structured output equally. These metrics draw inspiration from the currently used Symbol Error Rate and some ideas from Hajič jr. \cite{hajic_jr_further_2016}. We have divided our proposed metrics as tiers depending on the abstraction level they address and the problems they can help diagnose.
\subsection{Tier 0: Methodology-Specific Metrics}
The first tier is that devoted to specific pipeline metrics before performing any kind of reconstruction into the MTN format. The idea behind this tier is that comparison of methodology-specific metrics is still useful, since it might help identify strengths or shortcomings on certain applications and help quantify the effectiveness of methods converting into MTN format (which depending on the OMR methodology might be more or less nuanced).

Since the approaches to OMR are both diverse and unique, it does not make much sense to overly specify what metrics to employ under this tier, but rather use those metrics that are already being employed. Some examples of the kinds of metrics that should be reported here include (but are not restricted to):

\begin{itemize}
    \item For object detection methods, metrics such as mean average precision, F1 score, average intersection over union and so on.
    \item For end-to-end string-based methods, metrics such as symbol error rate or string parseability (whether or not the output string can be interpreted correctly with whichever grammar describes its underlying language).
\end{itemize}
\subsection{Tier 1: Primitive detection}
The first set of metrics addresses the presence or absence of terminals within the MTN string. The main goal is determining the capacity of models of detecting objects and analysing token co-ocurrence phenomena. This is especially important in OMR because the distribution of classes in music is extremely unbalanced: Noteheads are orders of magnitude more common than almost any other token. Understanding what tokens are missed and why is key to developing a good recognition system, since missing a single stem can alter the reconstruction of the score completely.

These metrics do not take into account structural matters, making them quick to compute. Given a set of unique terminals $T$, which is a subset of the full MTN vocabulary $V$, the multiset of predicted terminals within a sample $P = \left\{p_1 \dots p_n: p_i \in T \right\}$ as the multiset of all terminals produced by a model and the multiset of ground truth terminals within a sample $G = \left\{g_1 \dots g_m: g_i \in T \right\}$ as the multiset of all terminals existing in the groundtruth (both in no particular order), we define the following metrics:

\begin{itemize}
    \item Primitive-level precision
    \begin{equation}
        precision = \frac{\|P \cap G\|}{\|P\|}
    \end{equation}
    \item Primitive-level recall
    \begin{equation}
        recall = \frac{\|P \cap G\|}{\|G\|}
    \end{equation}
\end{itemize}
These metrics are computed per-class for the entire dataset. In order to produce a single precision and recall measure, results are aggregated per-class using a weighted average, where the weights are the relative frequency of each token in the ground truth.
\subsection{Tier 2: Structure Reconstruction}
This tier takes into account the structure of the produced MTN and compares it directly with that of the ground truth. A matching from ground truth elements to those present in the prediction is performed using a tree edit distance algorithm. In particular, since there is a restriction on the ordering of sibling labels, the $O\left(n^3\right)$ solution from Zhang and Sasha can be employed \cite{zhang_simple_1989}. In practice, we use a Python implementation \cite{JoaoFelipeAptedPython} of Pawlik \textit{et al.}'s APTED algorithm \cite{pawlikTreeEditDistance2016a}.

Given the following operations:
\begin{itemize}
    \item \textbf{Substitution:} Changing the label of a single node within the tree.
    \item \textbf{Deletion:} Removing a single node of the tree and setting its children as siblings.
    \item \textbf{Insertion:} Adding a new node under a parent one and setting a consecutive subsequence of its siblings as children.
\end{itemize}

Given a predicted tree and a ground truth tree whose set of vertices is $G$ and assuming an equal edit cost of 1 for all operations, the Tree Error Rate (TER) is defined as
\begin{equation}
    TER = \frac{S + D + I}{\|G\|}
\end{equation}
where $S$, $D$ and $I$ are the number of substitution, deletion and insertion operations required to produce the ground truth tree from the predicted tree. This metric is designed mostly for benchmarking and is defined by analogy to the ubiquitous Symbol Error Rate (SER).
\subsection{Tier 3: Semantic Reconstruction}
This tier considers whether the subset of music semantics required by MTN has been extracted correctly. It depends on the matching extracted from the structural level in order to identify the association between objects and their ground truth correspondences. Since evaluation has to be possible at the measure level, the graphical definition of pitch is used, which is the position of the notehead within the system.

A note $n$ is defined as a 3-tuple $(p, t, d) \in \mathbb{Z}^{2} \times \mathbb{Q} \times \mathbb{Q}$, where $p$ is a tuple of the staff the note belongs to and its position within the staff, $t$ is the number of beats from the start of the measure and $d$ is the duration of the note in number of beats. Rests are considered notes whose pitch is irrelevant (in practical terms, they can be thought of having a unique shared number for pitch). Given the set of predicted notes $P$ and the set of ground truth notes $G$, the matching between $P$ and $G$ is defined as a set of tuples $M = (n_p, n_g) \in P \times G$ such that at most one tuple with either a specific $n_p$ or $n_g$ exists.

The Missing Note Rate (MNR) is defined as the ratio of ground truth notes that do not have a corresponding prediction. More formally, 
\begin{equation}
    MNR = \frac{\| \left\{ n_g \in G : (n_p, n_g) \notin M, \forall n_p \in P \right\} \|}{\| G \|}.
\end{equation}
Similarly, the False Positive Rate (FPR) is defined as the ratio of predicted notes that do not have a corresponding ground truth note. It is defined by analogy to MNR.

Pitch Precision (PP) is defined as the number of correctly predicted pitches w.r.t. the ground truth as defined by the matching $M$. Therefore,
\begin{equation}
    PP = \frac{\| \left\{ (n_p, n_g) \in M : p_{n_p} = p_{n_g} \right\} \|}{\| M \|}.
\end{equation}
Time Precision (TP) is defined analogously.

The Average Pitch Shift (APS) is defined as the average offset in pitch from the predicted note w.r.t. its corresponding ground truth note as defined by the matching $M$. Therefore,
\begin{equation}
    APS = \frac{1}{\| M \|} \sum_{\forall (n_p, n_g) \in M} p_{n_p} - p_{n_g}.
\end{equation}
Time Average Shift (TAS) is defined analogously. Signedness is kept in order to identify the direction in which the underlying OMR system tends to move the notes.

In order for all of these metrics to be independent of the sequence length, they should be computed and accumulated for the entire dataset and not averaged on a by-prediction basis.

There is also the possibility of extending the TER metric with some semantics in order to ensure better matching and more fine-grained information from it. Instead of matching the predicted and ground truth trees with their structure alone, it would be interesting to incorporate pitch into the notehead elements. Thus, a pitch change can be though of a partial substitution of the node at a reduced cost of $0.5$ for either the staff, the position on the staff or the notehead type. If more than one of this properties change, it should be considered a full substitution.
\section{The COMREF Dataset} \label{sec:data}
We have developed a dataset (dubbed COMREF from ``Common Optical Music Recognition Evaluation Framework'')
built on transcriptions of public domain works as a proof of concept of the notation format. In particular, we have used the OpenScore project's transcriptions
of widely known works such as The Art of the Fugue by J.S. Bach or the Planets by Gustav Holst, among others. We
have also incorporated the Lieder Corpus \cite{GothamJonas2022} and the String Quartet corpus \cite{stringquartet_corpus}. All these scores are engraved from MusicXML files, but we plan on supporting MEI as a conversion source at some point in the future.
\subsection{Images}
In summary, the dataset is produced by processing of 894 individual works into images at the measure level (including all staves that belong to it), to produce a total of $435.623$ images. After verification, $461$ images are removed due to noise in the annotations, totalling $435.162$ images after cleanup.

\begin{figure*}
    \centering
    \includegraphics[width=\textwidth]{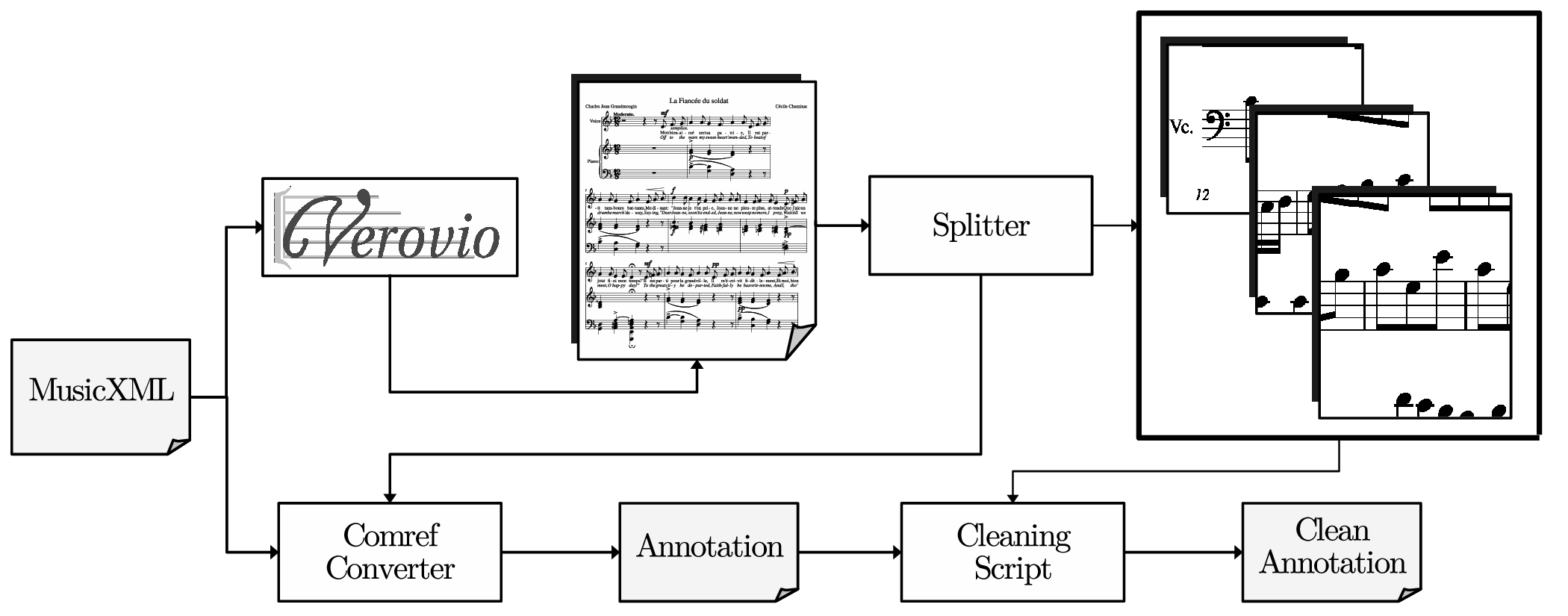}
    \caption{The pipeline through which the COMREF dataset has been generated.}
    \label{fig:generation}
\end{figure*}

The process through which the dataset was produced is summarised in Figure \ref{fig:generation}. Scores are engraved through Verovio \cite{Verovio}, a MEI-based score engraving system which is also compatible with MusicXML, into page-level SVG files. Using the hierarchical structure within the SVG and exploiting the optional identifier information Verovio can be instructed to attach, we employ a splitter script that finds bounding boxes for every measure and engraves them individually. It also finds what measures lie at the beginning of a line in order to properly add this information on the notation. The bounding boxes for all measures are cropped at the next staff vertically and a fixed amount of pixels horizontally.

Once the images are produced, the converter uses the MusicXML file and produces the MTN notation. In order to ensure all images have their corresponding ground truth, we use a cleaning script that finds matching identifiers for images in the MTN files. It also checks for outliers in case there are blatant mistakes in the notation. Although we have taken precautions to minimise the number of errors, there are a few images with objects far from the staff, either temporally or graphically. We remove these outliers heuristically to ensure the quality of the data.

\begin{figure*}
    \centering
    \includegraphics[width=.85\textwidth]{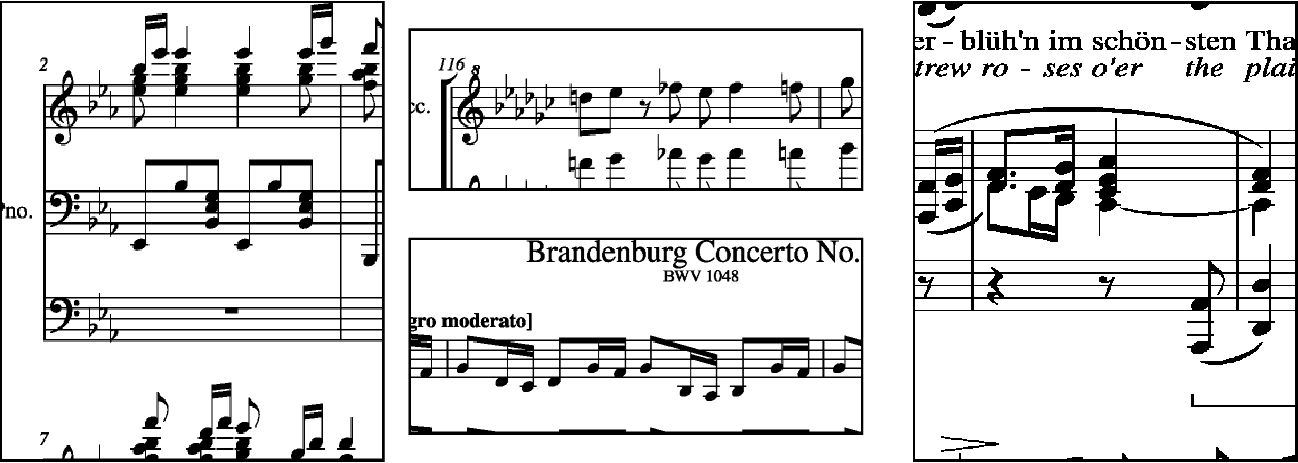}
    \caption{Some example measures sampled from the dataset. The dataset contains scores of varying types. On the left, an example of a pianoform score with three staves. In the middle, two examples of single-staff scores where the top one is part of a system of staves. On the right, a two-staff pianoform with two staves.}
    \label{fig:example_dataset}
\end{figure*}

Some qualitative examples of the generated samples can be seen in Figure \ref{fig:example_dataset}. The dataset purposefully crops images coarsely, incorporating parts of neighboring staves and surrounding distractors. This is in order to better emulate the result of a layout analysis algorithm, which will be inexact. The page-level images from which the measure-level dataset has been produced is also included for use cases that might require it, as well as the Verovio SVG output.

Overall, the resulting dataset is quite challenging because it incorporates a lot of variability, a selection of highly complex scores and lots of distractors.
\subsection{Ground Truth}
The dataset has 88 classes which are represented very unevenly. Noteheads, beams, stems and flags are the most common, adding up to 70\% of the entire dataset on their own. Of these, 71 are represented in a predefined test partition containing 52884 measures.

It includes music from a wide range of sources, including scores that are monophonic, polyphonic and pianoform. Diverse instrumentations and styles are also covered, with works for large orchestras coexisting with small-scope chamber music or solo works.

\subsection{Baseline Experiment}
We conducted a simple proof-of-concept experiment on the COMREF dataset to assess the feasibility of the methodology proposed in this paper. For this purpose, we used an off-the-shelf OMR system to produce a transcription of the test partition and we analysed its results.

The OMR system used for this experiment is Audiveris \cite{Audiveris}, an Open Source page-level system capable of generating a MusicXML output from a single input image. We used this model because it can generate a full score output in MusicXML, it can be run on the full batch of images directly and offers decent results. Other systems in the community or the literature either omit the semantic reconstruction step or cannot be run on the thousands of images of the test set efficiently. Implementing the score reconstruction step for low-level intermediate music representations -- e.g. bounding boxes, text representations -- is outside the scope of this work. Moreover, few models are designed to address multiple-staff polyphonic music.

The page-level images of the dataset are used as input, since Audiveris requires the information of the clef, key and beat. The output MusicXML is then converted to MTN and a simple matching between predicted and ground truth samples is generated by imposing a top-down reading order given the samples known to be present on each page. If a prediction has more measures per page than the ones in the ground truth, the extra ones are just discarded.

 With the setup outlined above, Audiveris predicted 45822 measures from the 52884 present on the ground truth. Out of these, 40622 measures from both sets could be matched together, corresponding to a coverage of 76.9\%. The missed predictions are as a result of the engine failing to give an output on certain pages.

Given this setup, the results for Tier 1 are shown in Table \ref{tab:tier1} and the results for Tiers 2 and 3 are shown in Table \ref{tab:tier23}.

\begin{table*}[ht]
\caption{Results for Tier 1 computation on the Dataset's test partition.}
\label{tab:tier1}
\begin{adjustbox}{max width=\textwidth}
\begin{tabular}{@{}lllllllllll@{}}
\cmidrule(r){1-5} \cmidrule(l){7-11}
Token                  & Precision & Recall & Counts & Prop   &  & Token                     & Precision & Recall & Counts & Prop   \\ \cmidrule(r){1-5} \cmidrule(l){7-11} 
notehead\_black        & 0.946     & 0.791  & 193879 & 0.2992 &  & barline\_tok\_heavy       & 0.812     & 0.913  & 643    & 0.0010 \\
stem\_down             & 0.925     & 0.816  & 99692  & 0.1539 &  & dyn\_pp                   & 0.901     & 0.714  & 597    & 0.0009 \\
stem\_up               & 0.896     & 0.756  & 78848  & 0.1217 &  & fermata                   & 0.960     & 0.577  & 589    & 0.0009 \\
beam                   & 0.915     & 0.769  & 48918  & 0.0755 &  & tuplet\_start             & 0.239     & 0.393  & 563    & 0.0009 \\
flag                   & 0.922     & 0.778  & 28190  & 0.0435 &  & dyn\_ff                   & 0.955     & 0.813  & 465    & 0.0007 \\
notehead\_white        & 0.582     & 0.715  & 20330  & 0.0314 &  & dyn\_mf                   & 0.891     & 0.742  & 430    & 0.0007 \\
accidental\_sharp      & 0.941     & 0.507  & 19882  & 0.0307 &  & trill                     & 0.884     & 0.637  & 408    & 0.0006 \\
slur\_start            & 0.852     & 0.685  & 17329  & 0.0267 &  & rest\_maxima              & 0.126     & 0.648  & 244    & 0.0004 \\
slur\_stop             & 0.850     & 0.675  & 17320  & 0.0267 &  & dyn\_sfz                  & 0.900     & 0.117  & 230    & 0.0004 \\
accidental\_flat       & 0.907     & 0.412  & 15350  & 0.0237 &  & wavy\_line                & 0.000     & 0.000  & 200    & 0.0003 \\
rest\_eighth           & 0.931     & 0.742  & 13821  & 0.0213 &  & notehead\_cue\_black      & 0.000     & 0.000  & 185    & 0.0003 \\
dot                    & 0.904     & 0.685  & 10894  & 0.0168 &  & rest\_32nd                & 0.957     & 0.817  & 164    & 0.0003 \\
staccato               & 0.637     & 0.641  & 9150   & 0.0141 &  & timesig\_common           & 0.464     & 0.634  & 131    & 0.0002 \\
accidental\_natural    & 0.926     & 0.738  & 8537   & 0.0132 &  & rest\_whole               & 0.082     & 0.047  & 106    & 0.0002 \\
rest\_quarter          & 0.920     & 0.794  & 8092   & 0.0125 &  & repeat\_backward          & 0.653     & 0.752  & 105    & 0.0002 \\
tied\_stop             & 0.822     & 0.496  & 7784   & 0.0120 &  & accidental\_double\_flat  & 1.000     & 0.077  & 104    & 0.0002 \\
tied\_start            & 0.812     & 0.466  & 7745   & 0.0120 &  & accidental\_double\_sharp & 1.000     & 0.646  & 96     & 0.0001 \\
clef\_G                & 0.882     & 0.354  & 5974   & 0.0092 &  & clef\_oct\_F              & 0.000     & 0.000  & 81     & 0.0001 \\
clef\_F                & 0.554     & 0.243  & 3120   & 0.0048 &  & dyn\_fz                   & 0.000     & 0.000  & 78     & 0.0001 \\
rest\_long             & 0.547     & 0.695  & 3039   & 0.0047 &  & dyn\_mp                   & 0.846     & 0.786  & 70     & 0.0001 \\
wedge\_stop            & 0.892     & 0.514  & 2849   & 0.0044 &  & clef\_oct\_G              & 0.279     & 0.185  & 65     & 0.0001 \\
rest\_breve            & 0.861     & 0.519  & 2761   & 0.0043 &  & repeat\_forward           & 0.295     & 0.316  & 57     & 0.0001 \\
accent                 & 0.802     & 0.677  & 2243   & 0.0035 &  & dyn\_sffz                 & 0.000     & 0.000  & 47     & 0.0001 \\
barline\_tok\_regular  & 0.787     & 0.899  & 2023   & 0.0031 &  & dyn\_ppp                  & 0.000     & 0.000  & 44     & 0.0001 \\
dyn\_p                 & 0.771     & 0.716  & 1707   & 0.0026 &  & timesig\_cut              & 0.818     & 0.878  & 41     & 0.0001 \\
dyn\_f                 & 0.867     & 0.698  & 1669   & 0.0026 &  & dyn\_fff                  & 0.000     & 0.000  & 39     & 0.0001 \\
arpeggiate             & 0.883     & 0.062  & 1573   & 0.0024 &  & dyn\_rf                   & 0.000     & 0.000  & 35     & 0.0001 \\
rest\_half             & 0.908     & 0.803  & 1488   & 0.0023 &  & dyn\_fp                   & 0.947     & 0.581  & 31     & 0.0000 \\
wedge\_diminuendo      & 0.874     & 0.641  & 1445   & 0.0022 &  & notehead\_breve           & 0.000     & 0.000  & 24     & 0.0000 \\
wedge\_crescendo       & 0.907     & 0.710  & 1433   & 0.0022 &  & notehead\_cue\_white      & 0.000     & 0.000  & 11     & 0.0000 \\
rest\_16th             & 0.936     & 0.819  & 1418   & 0.0022 &  & segno                     & 0.000     & 0.000  & 2      & 0.0000 \\
notehead\_grace\_black & 0.250     & 0.001  & 1297   & 0.0020 &  & dyn\_pppppp               & 0.000     & 0.000  & 1      & 0.0000 \\
clef\_C                & 0.423     & 0.442  & 954    & 0.0015 &  & coda                      & 0.000     & 0.000  & 1      & 0.0000 \\
dyn\_sf                & 0.609     & 0.671  & 671    & 0.0010 &  & turn                      & 0.083     & 1.000  & 1      & 0.0000 \\
tenuto                 & 0.457     & 0.673  & 651    & 0.0010 &  & caesura                   & 0.000     & 0.000  & 1      & 0.0000 \\
                       &           &        &        &        &  & \textbf{Total}                     & 0.894     & 0.733  & 647965 & 1.0      \\ \cmidrule(r){1-5} \cmidrule(l){7-11} 
\end{tabular}
\end{adjustbox}
\end{table*}
\begin{table*}[ht]
\caption{Results for Tiers 2 and 3. From left to right, Tree Error Rate, Average Time Shift, Average Pitch Shift, Time Precision, Pitch Precision, Staff Precision, False Positive Rate and Missing Note Rate}
\centering
\label{tab:tier23}
\begin{tabular}{@{}ccccccccc@{}}
\toprule
TER   & Time Shift     & Pitch Shift     & Staff Shift    & Time Prec.    & Pitch Prec.    & Staff Prec.    & FPR   & MNR   \\ \midrule
0.372 & -0.096 & -0.091 & 0.022 & 0.802 & 0.749 & 0.963 & 0.097 & 0.216 \\ \bottomrule
\end{tabular}
\end{table*}

From Tier 1, a few conclusions can be extracted. The first one is that Audiveris can find the structural music elements very reliably (noteheads, stems, beams and flags), all of them with recalls over 75\%. This is consistent with the values seen in the missing note rate at roughly 20\%, indicating that a majority of the detected noteheads is properly constructed and matched against noteheads in the ground truth. The system is also precise characterising the notes that can be matched, showing a pitch and time shift of roughly $-0.09$. This means that out of every 10 notes, only one is a single step below what it should.

The tree error rate is sensibly high, at $0.372$. This is attributed particularly to the missing elements. Upon inspection of certain samples, the most recurrent error is the omission of certain note groups, causing all of their attached elements to be missed.

The reason for the low scores in attribute elements (clefs, accidentals and time signature elements) is the fact that MusicXML does not encode start-of-line attribute elements because they are layout-dependent. By artificially inserting the tokens corresponding to start-of-sequence elements, we can adjust this to increase the recall but since the matching is sometimes imperfect we also insert a considerable amount of false positives.

Overall, even if the results for this specific tool on the dataset still leave room for improvement, we consider that our proposed format and metric fulfil their design purposes: unique representation of scores and evaluation. Therefore, we consider this simple trial successful.

\section{Conclusions} \label{sec:conclusion}
In this paper we have argued for the implementation of a Common Optical Music Recognition Framework through the instauration of a notation format in which score
reconstruction is independent from the recognition methodology. Moreover, the resulting scores can be evaluated fairly an unambiguously. Our proposed reification of this
idea is the MTN format, which draws heavy inspiration from existing formats and adapts
many ideas already present in the field while also being quite expressive and general.
Its main idea is exploiting the expressiveness of graphs while keeping the ordering properties of
sequences. Since this method builds upon some of the most widely used abstractions of the community (e.g. symbols as combinations of primitives, time from ordering, etc) it stands as a good candidate for a common endpoint for OMR as a whole.
Of course, CWMN is a tremendously complex notation system which has been optimised and
streamlined for hundreds of years. Nevertheless, we believe the subset of music that can be expressed
in this format is large enough to be useful for the community.

In this work, we have also presented a concrete implementation of a set of metrics for OMR
practitioners with the hopes of bringing together the community to speak the same
language; a \textit{lingua franca} thanks to which research can be shared and compared
fairly and easily. We provide a simple baseline from which to demonstrate how the evaluation framework works. We also hope this might lower the barrier of entry and bring some
new interest in the field.

The work that lies ahead now is building a corpus of music that can be employed with this
format into a benchmark for CWMN recognition, both in typeset and handwritten domains. 
For typeset scores, building a corpus in MTN is as straightforward as compiling
a comprehensive set of scores in MusicXML and engraving them. For handwritten music, the main challenge lies in developing techniques to reliably align MTN to the source material andfinding good transcriptions, which shall be the focus of our next efforts.
\backmatter

\bmhead{Supplementary information}

Link to the code: \url{https://github.com/CVC-DAG/comref-converter}.

\noindent Link to the dataset: \url{https://datasets.cvc.uab.cat/comref/comref.zip}.
\bmhead{Acknowledgments}
We gratefully thank the participation of Carles Badal and Jan Hajič Jr. in discussions that led to improvements on this paper.

This work has been partially supported by the Spanish projects PID2021-126808OB-I00 (GRAIL) and CNS2022-135947 (DOLORES). Pau Torras is funded by the Spanish FPU Grant FPU22/00207 and developed part of this work while funded by the AGAUR Joan Oró FI grant 2023 FI-1-00324. The authors acknowledge the support of the Generalitat de Catalunya CERCA Program to CVC’s general activities.

\section*{Declarations}

The authors have no conflicts of interest to declare that are relevant to the content of this article.





\bibliographystyle{bst/sn-basic.bst}
\bibliography{main}

\begin{thebibliography}{60}
\providecommand{\natexlab}[1]{#1}
\providecommand{\url}[1]{{#1}}
\providecommand{\urlprefix}{URL }
\providecommand{\doi}[1]{\url{https://doi.org/#1}}
\providecommand{\eprint}[2][]{\url{#2}}
 \bibcommenthead

\bibitem[{noa(2008)}]{noauthor_bach_nodate}
 (2008) Bach {Digital}.
  \urlprefix\url{https://www.bach-digital.de/content/index.xed}, accessed:
  2023-03-01

\bibitem[{Joa(2017)}]{JoaoFelipeAptedPython}
 (2017) {{JoaoFelipe}}/apted: {{Python APTED}} algorithm for the {{Tree Edit
  Distance}}. https://github.com/JoaoFelipe/apted/tree/master, accessed:
  2024-03-10

\bibitem[{noa(2023)}]{noauthor_beethoven-haus_nodate}
 (2023) Beethoven-{Haus} {Bonn}.
  \urlprefix\url{https://www.beethoven.de/en/archive/list}, accessed:
  2023-03-01

\bibitem[{str(2023)}]{stringquartet_corpus}
 (2023) String quartet corpus.
  \urlprefix\url{https://github.com/OpenScore/StringQuartets}, accessed:
  2023-10-10

\bibitem[{Alfaro-Contreras and
  Valero-Mas(2021)}]{alfaro-contreras_exploiting_2021}
Alfaro-Contreras M, Valero-Mas JJ (2021) Exploiting the two-dimensional nature
  of agnostic music notation for neural optical music recognition. Applied
  Sciences 11(8):3621. Publisher: MDPI

\bibitem[{Alfaro-Contreras et~al(2022)Alfaro-Contreras, Ríos-Vila, Valero-Mas,
  Iñesta, and Calvo-Zaragoza}]{alfaro-contreras_decoupling_2022}
Alfaro-Contreras M, Ríos-Vila A, Valero-Mas JJ, et~al (2022) Decoupling music
  notation to improve end-to-end {Optical} {Music} {Recognition}. Pattern
  Recognition Letters 158:157--163. \doi{10.1016/j.patrec.2022.04.032},
  \urlprefix\url{https://www.sciencedirect.com/science/article/pii/S0167865522001428}

\bibitem[{Baró et~al(2016)Baró, Riba, and Fornés}]{baro_towards_2016}
Baró A, Riba P, Fornés A (2016) Towards the {Recognition} of {Compound}
  {Music} {Notes} in {Handwritten} {Music} {Scores}. In: 2016 15th
  {International} {Conference} on {Frontiers} in {Handwriting} {Recognition}
  ({ICFHR}), pp 465--470, \doi{10.1109/ICFHR.2016.0092}, iSSN: 2167-6445

\bibitem[{Baró et~al(2019)Baró, Riba, Calvo-Zaragoza, and
  Fornés}]{baro_optical_2019}
Baró A, Riba P, Calvo-Zaragoza J, et~al (2019) From {Optical} {Music}
  {Recognition} to {Handwritten} {Music} {Recognition}: {A} baseline. Pattern
  Recognition Letters 123:1--8. \doi{10.1016/j.patrec.2019.02.029},
  \urlprefix\url{https://linkinghub.elsevier.com/retrieve/pii/S0167865518303386}

\bibitem[{Baró et~al(2020)Baró, Badal, and Fornés}]{baro_handwritten_2020}
Baró A, Badal C, Fornés A (2020) Handwritten {Historical} {Music}
  {Recognition} by {Sequence}-to-{Sequence} with {Attention} {Mechanism}. In:
  2020 17th {International} {Conference} on {Frontiers} in {Handwriting}
  {Recognition} ({ICFHR}), pp 205--210, \doi{10.1109/ICFHR2020.2020.00046}

\bibitem[{Baró et~al(2022)Baró, Riba, and Fornés}]{baro_musigraph_2022}
Baró A, Riba P, Fornés A (2022) Musigraph: {Optical} {Music} {Recognition}
  {Through} {Object} {Detection} and {Graph} {Neural} {Network}. In: Porwal U,
  Fornés A, Shafait F (eds) Frontiers in {Handwriting} {Recognition}. Springer
  International Publishing, Cham, Lecture {Notes} in {Computer} {Science}, pp
  171--184, \doi{10.1007/978-3-031-21648-0\_12}

\bibitem[{Bitteur(2004)}]{bitteur_audiveris_2004}
Bitteur H (2004) Audiveris. \urlprefix\url{https://github.com/audiveris}

\bibitem[{Bui et~al(2014)Bui, Na, and Kim}]{bui_staff_2014}
Bui HN, Na IS, Kim SH (2014) Staff {Line} {Removal} {Using} {Line} {Adjacency}
  {Graph} and {Staff} {Line} {Skeleton} for {Camera}-{Based} {Printed} {Music}
  {Scores}. In: 22nd {International} {Conference} on {Pattern} {Recognition},
  pp 2787--2789, \doi{10.1109/ICPR.2014.480}, iSSN: 1051-4651

\bibitem[{Byrd and Simonsen(2015)}]{byrd_towards_2015}
Byrd D, Simonsen JG (2015) Towards a {Standard} {Testbed} for {Optical} {Music}
  {Recognition}: {Definitions}, {Metrics}, and {Page} {Images}. Journal of New
  Music Research 44(3):169--195. \doi{10.1080/09298215.2015.1045424},
  \urlprefix\url{http://www.tandfonline.com/doi/full/10.1080/09298215.2015.1045424}

\bibitem[{Byrd and Isaacson(2003)}]{byrd_music_2003}
Byrd DA, Isaacson EJ (2003) A {Music} {Representation} {Requirement}
  {Specification} for {Academia}. Computer Music Journal 27(4):43--57.
  \urlprefix\url{https://muse.jhu.edu/pub/6/article/49604}, publisher: The MIT
  Press

\bibitem[{Calvo-Zaragoza and Oncina(2014)}]{calvo-zaragoza_recognition_2014}
Calvo-Zaragoza J, Oncina J (2014) Recognition of {Pen}-{Based} {Music}
  {Notation}: {The} {HOMUS} {Dataset}. In: 22nd {International} {Conference} on
  {Pattern} {Recognition}. Institute of Electrical \& Electronics Engineers
  (IEEE), pp 3038--3043, \doi{10.1109/ICPR.2014.524}, iSSN: 1051-4651

\bibitem[{Calvo-Zaragoza and
  Rizo(2018{\natexlab{a}})}]{calvo-zaragoza_camera-primus_2018}
Calvo-Zaragoza J, Rizo D (2018{\natexlab{a}}) Camera-{PrIMuS}: {Neural}
  {End}-to-{End} {Optical} {Music} {Recognition} on {Realistic} {Monophonic}
  {Scores}. In: 19th {International} {Society} for {Music} {Information}
  {Retrieval} {Conference}, Paris, France, pp 248--255,
  \urlprefix\url{http://ismir2018.ircam.fr/doc/pdfs/33_Paper.pdf}

\bibitem[{Calvo-Zaragoza and
  Rizo(2018{\natexlab{b}})}]{calvo-zaragoza_end--end_2018}
Calvo-Zaragoza J, Rizo D (2018{\natexlab{b}}) End-to-{End} {Neural} {Optical}
  {Music} {Recognition} of {Monophonic} {Scores}. Applied Sciences 8(4):606.
  \doi{10.3390/app8040606},
  \urlprefix\url{https://www.mdpi.com/2076-3417/8/4/606}, number: 4 Publisher:
  Multidisciplinary Digital Publishing Institute

\bibitem[{Calvo-Zaragoza et~al(2017{\natexlab{a}})Calvo-Zaragoza, Pertusa, and
  Oncina}]{calvo-zaragoza_staff-line_2017}
Calvo-Zaragoza J, Pertusa A, Oncina J (2017{\natexlab{a}}) Staff-line detection
  and removal using a convolutional neural network. Machine Vision and
  Applications pp 1--10. \doi{10.1007/s00138-017-0844-4}

\bibitem[{Calvo-Zaragoza et~al(2017{\natexlab{b}})Calvo-Zaragoza, Valero-Mas,
  and Pertusa}]{calvo-zaragoza_end--end_2017}
Calvo-Zaragoza J, Valero-Mas JJ, Pertusa A (2017{\natexlab{b}}) End-to-{End}
  {Optical} {Music} {Recognition} {Using} {Neural} {Networks}. Suzhou, China,
  \doi{10.5281/zenodo.1418333},
  \urlprefix\url{https://zenodo.org/record/1418333}, iSBN: 9789811151798 Pages:
  472-477 Publisher: ISMIR Publication Title: Proceedings of the 18th
  International Society for Music Information Retrieval Conference

\bibitem[{Calvo-Zaragoza et~al(2021)Calvo-Zaragoza, Hajič~Jr., and
  Pacha}]{calvo-zaragoza_understanding_2021}
Calvo-Zaragoza J, Hajič~Jr. J, Pacha A (2021) Understanding {Optical} {Music}
  {Recognition}. ACM Comput Surv 53(4):1--35. \doi{10.1145/3397499},
  \urlprefix\url{https://dl.acm.org/doi/10.1145/3397499}

\bibitem[{Cardoso et~al(2009)Cardoso, Capela, Rebelo, Guedes, and Pinto~da
  Costa}]{cardoso_staff_2009}
Cardoso JdS, Capela A, Rebelo A, et~al (2009) Staff {Detection} with {Stable}
  {Paths}. IEEE Transactions on Pattern Analysis and Machine Intelligence
  31(6):1134--1139. \doi{10.1109/TPAMI.2009.34}, publisher: Institute of
  Electrical \& Electronics Engineers (IEEE)

\bibitem[{Edirisooriya et~al(2021)Edirisooriya, Dong, McAuley, and
  {Berg-Kirkpatrick}}]{edirisooriyaEmpiricalEvaluationEndtoEnd2021}
Edirisooriya S, Dong HW, McAuley J, et~al (2021) An {{Empirical Evaluation}} of
  {{End-to-End Polyphonic Optical Music Recognition}}. \eprint{2108.01769}

\bibitem[{Egozy and Clester(2022)}]{egozy_computer-assisted_2022}
Egozy E, Clester I (2022) Computer-{Assisted} {Measure} {Detection} in a
  {Music} {Score}-{Following} {Application}. In: Calvo-Zaragoza J, Pacha A,
  Shatri E (eds) Proceedings of the 4th {International} {Workshop} on {Reading}
  {Music} {Systems}, Online, pp 33--36, \doi{10.48550/arXiv.2211.13285},
  \urlprefix\url{https://sites.google.com/view/worms2022/proceedings}

\bibitem[{Fornés et~al(2013)Fornés, Dutta, Gordo, and
  Lladós}]{fornes_2012_2013}
Fornés A, Dutta A, Gordo A, et~al (2013) The 2012 {Music} {Scores}
  {Competitions}: {Staff} {Removal} and {Writer} {Identification}. In: Kwon YB,
  Ogier JM (eds) Graphics {Recognition}. {New} {Trends} and {Challenges}.
  Springer Berlin Heidelberg, Berlin, Heidelberg, pp 173--186,
  \doi{10.1007/978-3-642-36824-0_17}

\bibitem[{Foscarin et~al(2019)Foscarin, Jacquemard, and
  {Fournier-S'niehotta}}]{foscarinDiffProcedureMusic2019}
Foscarin F, Jacquemard F, {Fournier-S'niehotta} R (2019) A diff procedure for
  music score files. In: Proceedings of the 6th {{International Conference}} on
  {{Digital Libraries}} for {{Musicology}}. Association for Computing
  Machinery, New York, NY, USA, {{DLfM}} '19, pp 58--64,
  \doi{10.1145/3358664.3358671}

\bibitem[{Fujinaga(2004)}]{fujinaga_staff_2004}
Fujinaga I (2004) Staff detection and removal. In: Visual {Perception} of
  {Music} {Notation}: {On}-{Line} and {Off} {Line} {Recognition}. IGI Global, p
  1--39, \doi{10.4018/978-1-59140-298-5.ch001}

\bibitem[{Garrido-Munoz et~al(2022)Garrido-Munoz, Rios-Vila, and
  Calvo-Zaragoza}]{garrido-munoz_holistic_2022}
Garrido-Munoz C, Rios-Vila A, Calvo-Zaragoza J (2022) A holistic approach for
  image-to-graph: application to optical music recognition. IJDAR
  \doi{10.1007/s10032-022-00417-4},
  \urlprefix\url{https://doi.org/10.1007/s10032-022-00417-4}

\bibitem[{Good(2001)}]{good_musicxml_2001}
Good M (2001) {MusicXML}: {An} {Internet}-{Friendly} {Format} for {Sheet}
  {Music}. Tech. rep., Recordare LLC,
  \urlprefix\url{https://pdfs.semanticscholar.org/5617/972667ff794da79a4cbb6b985e85f8487ddd.pdf}

\bibitem[{Gotham and Jonas(2022)}]{GothamJonas2022}
Gotham MRH, Jonas P (2022) {The OpenScore Lieder Corpus}. In: M{\"u}nnich S,
  Rizo D (eds) {Music Encoding Conference Proceedings 2021}. {Humanities
  Commons}, pp 131--136, \doi{10.17613/1my2-dm23}

\bibitem[{Hajič and Pecina(2017)}]{hajic_muscima_2017}
Hajič J, Pecina P (2017) The {MUSCIMA}++ {Dataset} for {Handwritten} {Optical}
  {Music} {Recognition}. In: 2017 14th {IAPR} {International} {Conference} on
  {Document} {Analysis} and {Recognition} ({ICDAR}), pp 39--46,
  \doi{10.1109/ICDAR.2017.16}, iSSN: 2379-2140

\bibitem[{Hajič~jr.(2018)}]{hajic_jr_case_2018}
Hajič~jr. J (2018) A {Case} for {Intrinsic} {Evaluation} of {Optical} {Music}
  {Recognition}. In: Calvo-Zaragoza J, Hajič~jr. J, Pacha A (eds) 1st
  {International} {Workshop} on {Reading} {Music} {Systems}, Paris, France, pp
  15--16, \urlprefix\url{https://sites.google.com/view/worms2018/proceedings}

\bibitem[{Hajič~jr. et~al(2016)Hajič~jr., Novotný, Pecina, and
  Pokorný}]{hajic_jr_further_2016}
Hajič~jr. J, Novotný J, Pecina P, et~al (2016) Further {Steps} towards a
  {Standard} {Testbed} for {Optical} {Music} {Recognition}. In: Mandel M,
  Devaney J, Turnbull D, et~al (eds) 17th {International} {Society} for {Music}
  {Information} {Retrieval} {Conference}. New York University, New York, USA,
  pp 157--163, \urlprefix\url{https://wp.nyu.edu/ismir2016/event/proceedings/},
  backup Publisher: New York University

\bibitem[{Huang et~al(2019)Huang, Jia, and Guo}]{huang_state---art_2019}
Huang Z, Jia X, Guo Y (2019) State-of-the-{Art} {Model} for {Music} {Object}
  {Recognition} with {Deep} {Learning}. Applied Sciences 9(13):2645--2665.
  \doi{10.3390/app9132645},
  \urlprefix\url{https://www.mdpi.com/2076-3417/9/13/2645}

\bibitem[{Huron(20??)}]{HumdrumToolkitComputational}
Huron D (20??) The {{Humdrum Toolkit}} for {{Computational Music Analysis}}
  {\textbar} {{Humdrum}}. https://www.humdrum.org/, accessed: 2024-03-14

\bibitem[{Mengarelli et~al(2019)Mengarelli, Kostiuk, Vitório, Tibola, Wolff,
  and Silla}]{mengarelli_omr_2019}
Mengarelli L, Kostiuk B, Vitório JG, et~al (2019) {OMR} metrics and
  evaluation: a systematic review. Multimedia Tools and Applications
  \doi{10.1007/s11042-019-08200-0}

\bibitem[{Pacha(2017)}]{pacha_omr_2017}
Pacha A (2017) The {OMR} {Datasets} {Project}.
  \urlprefix\url{https://apacha.github.io/OMR-Datasets}

\bibitem[{Pacha(2018)}]{pacha_advancing_2018}
Pacha A (2018) Advancing {OMR} as a {Community}: {Best} {Practices} for
  {Reproducible} {Research}. In: Calvo-Zaragoza J, Hajič~jr. J, Pacha A (eds)
  1st {International} {Workshop} on {Reading} {Music} {Systems}, Paris, France,
  pp 19--20,
  \urlprefix\url{https://sites.google.com/view/worms2018/proceedings}

\bibitem[{Pacha(2021)}]{pacha_challenge_2021}
Pacha A (2021) The {Challenge} of {Reconstructing} {Digits} in {Music}
  {Scores}. In: Calvo-Zaragoza J, Pacha A (eds) Proceedings of the 3rd
  {International} {Workshop} on {Reading} {Music} {Systems}, Alicante, Spain,
  pp 4--7, \urlprefix\url{https://sites.google.com/view/worms2021/proceedings}

\bibitem[{Pacha et~al(2018)Pacha, Hajič~jr., and
  Calvo-Zaragoza}]{pacha_baseline_2018}
Pacha A, Hajič~jr. J, Calvo-Zaragoza J (2018) A {Baseline} for {General}
  {Music} {Object} {Detection} with {Deep} {Learning}. Applied Sciences
  8(9):1488--1508. \doi{10.3390/app8091488},
  \urlprefix\url{http://www.mdpi.com/2076-3417/8/9/1488}

\bibitem[{Parada-Cabaleiro et~al(2017)Parada-Cabaleiro, Batliner, Baird, and
  Schuller}]{parada-cabaleiro_seils_2017}
Parada-Cabaleiro E, Batliner A, Baird A, et~al (2017) The {SEILS} {Dataset}:
  {Symbolically} {Encoded} {Scores} in {Modern}-{Early} {Notation} for
  {Computational} {Musicology}. In: 18th {International} {Society} for {Music}
  {Information} {Retrieval} {Conference}, Suzhou, China,
  \urlprefix\url{https://ismir2017.smcnus.org/wp-content/uploads/2017/10/14_Paper.pdf}

\bibitem[{Pawlik and Augsten(2016)}]{pawlikTreeEditDistance2016a}
Pawlik M, Augsten N (2016) Tree edit distance: {{Robust}} and memory-efficient.
  Information Systems 56:157--173. \doi{10.1016/j.is.2015.08.004}

\bibitem[{Project(20??)}]{Audiveris}
Project A (20??) Audiveris - open-source optical music recognition.
  https://github.com/Audiveris/audiveris/, accessed: 2024-03-14

\bibitem[{Pugin(20??)}]{Verovio}
Pugin L (20??) Verovio, a music notation engraving library.
  https://www.verovio.org/, accessed: 2024-03-14

\bibitem[{Rebelo et~al(2012)Rebelo, Fujinaga, Paszkiewicz, Marcal, Guedes, and
  Cardoso}]{rebelo_optical_2012}
Rebelo A, Fujinaga I, Paszkiewicz F, et~al (2012) Optical music recognition:
  state-of-the-art and open issues. Int J Multimed Info Retr 1(3):173--190.
  \doi{10.1007/s13735-012-0004-6},
  \urlprefix\url{http://link.springer.com/10.1007/s13735-012-0004-6}

\bibitem[{Roland(2002)}]{roland_music_2002}
Roland P (2002) The music encoding initiative ({MEI}). In: 1st {International}
  {Conference} on {Musical} {Applications} {Using} {XML}, pp 55--59,
  \urlprefix\url{https://pdfs.semanticscholar.org/7fc4/16754b0508837dde8b505b3fd4dc517c7292.pdf}

\bibitem[{Ríos-Vila et~al(2022{\natexlab{a}})Ríos-Vila, Iñesta, and
  Calvo-Zaragoza}]{rios-vila_end--end_2022}
Ríos-Vila A, Iñesta JM, Calvo-Zaragoza J (2022{\natexlab{a}}) End-{To}-{End}
  {Full}-{Page} {Optical} {Music} {Recognition} of {Monophonic} {Documents} via
  {Score} {Unfolding}. In: Calvo-Zaragoza J, Pacha A, Shatri E (eds)
  Proceedings of the 4th {International} {Workshop} on {Reading} {Music}
  {Systems}, Online, pp 20--24, \doi{10.48550/arXiv.2211.13285},
  \urlprefix\url{https://sites.google.com/view/worms2022/proceedings}

\bibitem[{Ríos-Vila et~al(2022{\natexlab{b}})Ríos-Vila, Iñesta, and
  Calvo-Zaragoza}]{rios-vila_use_2022}
Ríos-Vila A, Iñesta JM, Calvo-Zaragoza J (2022{\natexlab{b}}) On the {Use}
  of {Transformers} for {End}-to-{End} {Optical} {Music} {Recognition}. In:
  Pinho AJ, Georgieva P, Teixeira LF, et~al (eds) Pattern {Recognition} and
  {Image} {Analysis}. Springer International Publishing, Cham, Lecture {Notes}
  in {Computer} {Science}, pp 470--481, \doi{10.1007/978-3-031-04881-4\_37}

\bibitem[{Ríos-Vila et~al(2023)Ríos-Vila, Rizo, Iñesta, and
  Calvo-Zaragoza}]{rios-vila_end--end_2023}
Ríos-Vila A, Rizo D, Iñesta JM, et~al (2023) End-to-end optical music
  recognition for pianoform sheet music. International Journal on Document
  Analysis and Recognition (IJDAR) 26(3):347--362.
  \doi{10.1007/s10032-023-00432-z},
  \urlprefix\url{https://doi.org/10.1007/s10032-023-00432-z}

\bibitem[{Shatri and Fazekas(2021)}]{shatri_doremi_2021}
Shatri E, Fazekas G (2021) {DoReMi}: {First} glance at a universal {OMR}
  dataset. In: Calvo-Zaragoza J, Pacha A (eds) Proceedings of the 3rd
  {International} {Workshop} on {Reading} {Music} {Systems}, Alicante, Spain,
  pp 43--49,
  \urlprefix\url{https://sites.google.com/view/worms2021/proceedings}

\bibitem[{Shishido et~al(2021)Shishido, Fati, Tokushige, and
  Ono}]{shishido_listen_2021}
Shishido T, Fati F, Tokushige D, et~al (2021) Listen to {Your} {Favorite}
  {Melodies} with {img2Mxml}, {Producing} {MusicXML} from {Sheet} {Music}
  {Image} by {Measure}-based {Multimodal} {Deep} {Learning}-driven {Assembly}.
  \urlprefix\url{http://arxiv.org/abs/2106.12037}, arXiv:2106.12037 [cs]

\bibitem[{{The LilyPond Developement
  Team}(2014)}]{the_lilypond_developement_team_lilypond_2014}
{The LilyPond Developement Team} (2014) {LilyPond} - {Essay} on automated music
  engraving. \urlprefix\url{http://www.lilypond.org/}

\bibitem[{Torras et~al(2021)Torras, Baró, Kang, and
  Fornés}]{torras_integration_2021}
Torras P, Baró A, Kang L, et~al (2021) On the {Integration} of {Language}
  {Models} into {Sequence} to {Sequence} {Architectures} for {Handwritten}
  {Music} {Recognition}. Online, \doi{10.5281/zenodo.5624451},
  \urlprefix\url{https://zenodo.org/record/5624451}, pages: 690-696 Publication
  Title: Proceedings of the 22nd International Society for Music Information
  Retrieval Conference Publisher: ISMIR

\bibitem[{Tuggener et~al(2018)Tuggener, Elezi, Schmidhuber, Pelillo, and
  Stadelmann}]{tuggener_deepscores_2018}
Tuggener L, Elezi I, Schmidhuber J, et~al (2018) {DeepScores} - {A} {Dataset}
  for {Segmentation}, {Detection} and {Classification} of {Tiny} {Objects}. In:
  24th {International} {Conference} on {Pattern} {Recognition}. ZHAW, Beijing,
  China, \doi{10.21256/zhaw-4255},
  \urlprefix\url{https://arxiv.org/abs/1804.00525}

\bibitem[{Tuggener et~al(2020)Tuggener, Satyawan, Pacha, Schmidhuber, and
  Stadelmann}]{tuggener_deepscoresv2_2020}
Tuggener L, Satyawan YP, Pacha A, et~al (2020) The {DeepScoresV2} {Dataset} and
  {Benchmark} for {Music} {Object} {Detection}. In: Proceedings of the 25th
  {International} {Conference} on {Pattern} {Recognition}, Milan, Italy,
  \doi{10.21256/zhaw-20647}

\bibitem[{Tuggener et~al(2023)Tuggener, Emberger, Ghosh, Sager, Satyawan,
  Montoya, Goldschagg, Seibold, Gut, Ackermann, Schmidhuber, and
  Stadelmann}]{tuggener_real_2023}
Tuggener L, Emberger R, Ghosh A, et~al (2023) Real world music object
  recognition. Transactions of the International Society for Music Information
  Retrieval \doi{10.21256/zhaw-28644},
  \urlprefix\url{https://digitalcollection.zhaw.ch/handle/11475/28644},
  accepted: 2023-09-08T13:52:37Z Publisher: Ubiquity Press

\bibitem[{Wen and Zhu(2022)}]{wen_sequence--sequence_2022}
Wen C, Zhu L (2022) A {Sequence}-to-{Sequence} {Framework} {Based} on
  {Transformer} {With} {Masked} {Language} {Model} for {Optical} {Music}
  {Recognition}. IEEE Access 10:118,243--118,252.
  \doi{10.1109/ACCESS.2022.3220878}, conference Name: IEEE Access

\bibitem[{Yu et~al(2020)Yu, Han, Gong, Jan, Zhao, Ye, Chen, Feng, Zhang, Wang,
  Xin, Liu, Mao, Xu, Zhang, Han, Gao, Tang, Jin, Hong, Yang, Li, Luo, Zhao, and
  Shi}]{yu_1st_2020}
Yu X, Han Z, Gong Y, et~al (2020) The 1st {Tiny} {Object} {Detection}
  {Challenge}: {Methods} and {Results}. In: Bartoli A, Fusiello A (eds)
  Computer {Vision} – {ECCV} 2020 {Workshops}. Springer International
  Publishing, Cham, Lecture {Notes} in {Computer} {Science}, pp 315--323,
  \doi{10.1007/978-3-030-68238-5_23}

\bibitem[{Zhang et~al(2018)Zhang, Du, and Dai}]{zhang_multi-scale_2018}
Zhang J, Du J, Dai L (2018) Multi-{Scale} {Attention} with {Dense} {Encoder}
  for {Handwritten} {Mathematical} {Expression} {Recognition}. In: 2018 24th
  {International} {Conference} on {Pattern} {Recognition} ({ICPR}), pp
  2245--2250, \doi{10.1109/ICPR.2018.8546031}, iSSN: 1051-4651

\bibitem[{Zhang and Shasha(1989)}]{zhang_simple_1989}
Zhang K, Shasha D (1989) Simple {Fast} {Algorithms} for the {Editing}
  {Distance} between {Trees} and {Related} {Problems}. SIAM J Comput
  18(6):1245--1262. \doi{10.1137/0218082},
  \urlprefix\url{https://epubs.siam.org/doi/abs/10.1137/0218082}, publisher:
  Society for Industrial and Applied Mathematics

\bibitem[{Zhang et~al(2023)Zhang, Huang, Zhang, and Ren}]{zhang_detector_2023}
Zhang Y, Huang Z, Zhang Y, et~al (2023) A detector for page-level handwritten
  music object recognition based on deep learning. Neural Comput \& Applic
  \doi{10.1007/s00521-023-08216-6},
  \urlprefix\url{https://doi.org/10.1007/s00521-023-08216-6}

\end{thebibliography}

\end{document}